\documentclass{bmvc2k}
\usepackage{mathcomp}
\usepackage{amsmath}
\usepackage{bm}
\usepackage{comment}
\usepackage{enumitem}

\title{DAOVI: Distortion-Aware Omnidirectional Video Inpainting}

\addauthor{Ryosuke Seshimo}{seshimo.ryosuke@keio.jp}{1}
\addauthor{Mariko Isogawa}{mariko.isogawa@keio.jp}{1}

\addinstitution{
 Keio University, Japan
}

\runninghead{SESHIMO ET AL.}{DAOVI}

\def\etal{\emph{et al}\bmvaOneDot}

\begin{document}

\maketitle

\begin{abstract}
Omnidirectional videos that capture the entire surroundings are employed in a variety of fields such as VR applications and remote sensing.
However, their wide field of view often causes unwanted objects to appear in the videos.
This problem can be addressed by video inpainting, which enables the natural removal of such objects while preserving both spatial and temporal consistency.
Nevertheless, most existing methods assume processing ordinary videos with a narrow field of view and do not tackle the distortion in equirectangular projection of omnidirectional videos.
To address this issue, this paper proposes a novel deep learning model for omnidirectional video inpainting, called Distortion-Aware Omnidirectional Video Inpainting (DAOVI).
DAOVI introduces a module that evaluates temporal motion information in the image space considering geodesic distance, as well as a depth-aware feature propagation module in the feature space that is designed to address the geometric distortion inherent to omnidirectional videos.
The experimental results demonstrate that our proposed method outperforms existing methods both quantitatively and qualitatively.
\end{abstract}

\begin{figure}[htbp]
  \centering
  \setlength{\tabcolsep}{1pt}
  \begin{tabular}{@{}ccccc@{}}
    \bmvaHangBox{\includegraphics[width=0.19\linewidth]{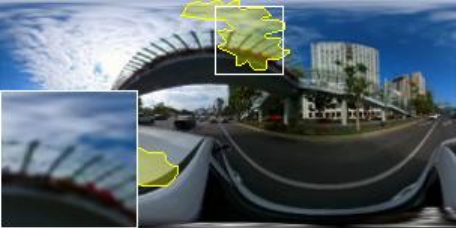}} &
    \bmvaHangBox{\includegraphics[width=0.19\linewidth]{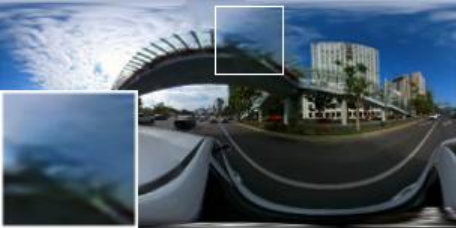}} &
    \bmvaHangBox{\includegraphics[width=0.19\linewidth]{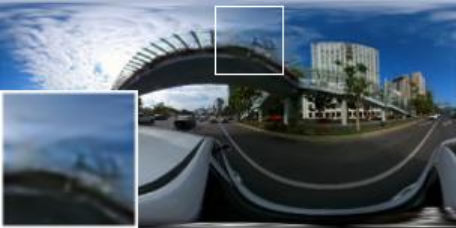}} &
    \bmvaHangBox{\includegraphics[width=0.19\linewidth]{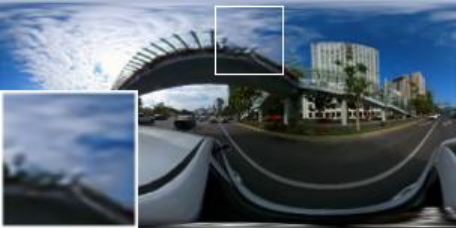}} &
    \bmvaHangBox{\includegraphics[width=0.19\linewidth]{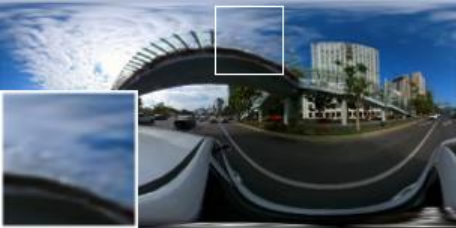}} \\

    \bmvaHangBox{\includegraphics[width=0.19\linewidth]{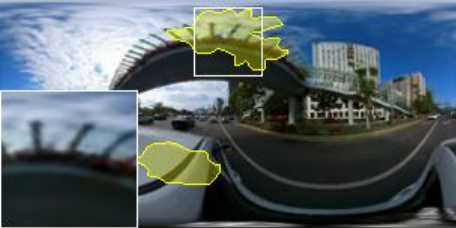}} &
    \bmvaHangBox{\includegraphics[width=0.19\linewidth]{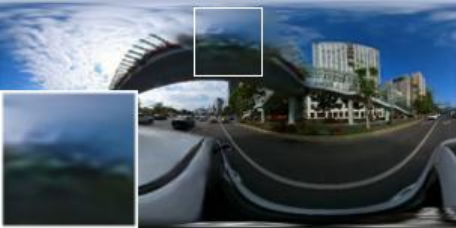}} &
    \bmvaHangBox{\includegraphics[width=0.19\linewidth]{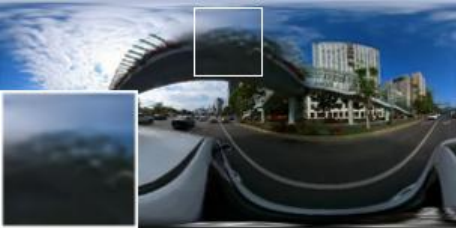}} &
    \bmvaHangBox{\includegraphics[width=0.19\linewidth]{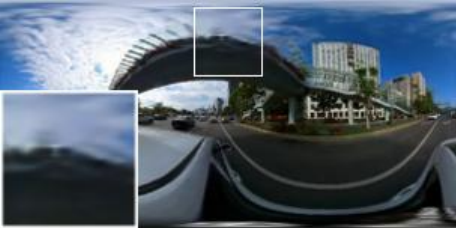}} &
    \bmvaHangBox{\includegraphics[width=0.19\linewidth]{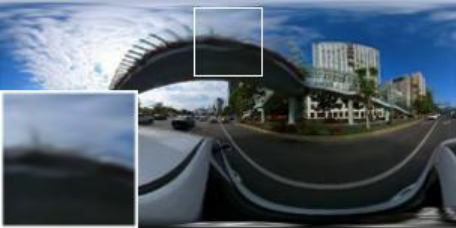}} \\
    Masked Frames & FuseFormer & STTN & ProPainter & DAOVI (Ours)
  \end{tabular}
  \caption{Inpainting results of the proposed DAOVI and several state-of-the-art video inpainting methods~\cite{FuseFormer, STTN, ProPainter} originally designed for ordinary videos with a narrow field of view. The yellow regions indicate the masked areas, and the outputs of each model within these regions are enlarged and shown at the bottom left. By accounting for geometric distortion in omnidirectional videos, our method produces more visually plausible results.}
  \label{fig:first_image}
\end{figure}

\section{Introduction}
\label{sec:intro}
Omnidirectional videos that capture the entire scene from every direction can provide users with an immersive experience and attract attention for use in VR/AR applications and remote sensing~\cite{ODV360, remote_sensing}.
Unlike ordinary videos with a narrow field of view (FoV), omnidirectional videos can represent the entire surroundings.
As a result, camera operators need not worry much about adjusting the camera angle to keep the subject in frame.
On the other hand, the wide FoV often causes unwanted objects to appear in the videos.

Video inpainting is a well-known technique for removing undesired regions inadvertently captured in videos.
This technique designates the regions to be removed as masked regions and reconstructs them seamlessly to match the surrounding content.
In particular, deep learning-based approaches~\cite{Proposal_based, ProPainter} have emerged in recent years, demonstrating their capability to achieve high-quality completions.

Many of the recently effective approaches incorporate optical flow based propagation~\cite{E2FGVI, ProPainter}.
In these methods, the masked regions in the target frame are filled by propagating information from reference frames.
This enables natural completion while maintaining spatial and temporal consistency.
Moreover, considering the influence of optical flow estimation errors in masked regions on the inpainting results, a depth-guided video inpainting method~\cite{DGDVI} has also been proposed.
This method uses depth estimation that is less prone to errors in masked regions to guide the inpainting process.
However, despite these advances, most existing video inpainting methods~\cite{DFVI, FGT, E2FGVI, ProPainter, STTN, DGDVI, FuseFormer} are designed for ordinary videos with a narrow FoV and are not suitable for omnidirectional videos, which include significant distortion due to equirectangular projection (ERP).
Fig.~\ref{fig:first_image} shows that applying these video inpainting methods to omnidirectional videos introduces artifacts, making it difficult to obtain plausible inpainting results.

Although methods have been proposed for omnidirectional video inpainting that account for distortion, many of these approaches~\cite{2009kawai, 2014kawai, 2016ROBIO} impose strict requirements for input videos, such as requiring a static background or planar masked regions.
As a result, it is challenging to apply these methods in a variety of scenarios.

To address this issue, this paper proposes a deep learning-based omnidirectional video inpainting method, called Distortion-Aware Omnidirectional Video Inpainting (DAOVI).
DAOVI consists of two main modules tailored for omnidirectional videos.
First, we introduce the Geodesic Flow-Consistent Image Propagation (GFCIP) module.
Conventional flow-based methods~\cite{DFVI, FGVC, ProPainter} compute the sum of the forward and backward optical flows at each pixel and evaluate flow validity by checking whether the sum is below a threshold.
Only valid flows are used to propagate pixel values in the image space.
However, distortion in omnidirectional videos varies with position, so using a single threshold is no longer reasonable.
Instead, GFCIP evaluates flow validity by checking the geodesic distance on the unit sphere.

Second, we propose the Omnidirectional Depth-Assisted Feature Propagation (ODAFP) module.
Conventional approaches~\cite{E2FGVI, ProPainter} propagate features in the latent space from adjacent frames using deformable convolutional networks (DCN)~\cite{DCNv1, DCNv2}, but they do not account for ERP distortion.
ODAFP generates DCN offsets and modulation masks tailored to omnidirectional videos by using convolutions and padding schemes designed for $360\tcdegree$ images and a distortion map that indicates the amount of ERP distortion at each pixel.
Furthermore, inspired by the finding that depth guidance can significantly enhance video inpainting~\cite{DGDVI}, depth maps are used as additional input to complement flow based propagation.

In summary, we make the following contributions:
\begin{itemize}[noitemsep, topsep=0pt, left=0pt]
    \item We propose a deep learning-based video inpainting method for omnidirectional videos, called Distortion-Aware Omnidirectional Video Inpainting (DAOVI).
    \item In the image space, we introduce the Geodesic Flow-Consistent Image Propagation (GFCIP) module, which evaluates optical flow validity using geodesic distance. 
    \item In the feature space, we introduce the Omnidirectional Depth-Assisted Feature Propagation (ODAFP) module that performs propagation using distortion guided modulation and convolutions designed for omnidirectional images, and uses depth maps as additional input to complement optical flow.
\end{itemize}

\section{Related Work}
\noindent \textbf{Video Inpainting.} 
Video inpainting is a technique for removing specified regions from a video and seamlessly restoring them.  
Various methods have been proposed in recent years~\cite{STTN,FuseFormer,CPNet,free_form_video_inpainting,Proposal_based, video_inpainting_by_jointly}.
Among them, many recent approaches utilize optical flow~\cite{DFVI,E2FGVI,ProPainter,FGVC,ISVI,FGT}.
For example, Xu \etal introduced an optical flow based video inpainting method called DFVI~\cite{DFVI}.
In this approach, rather than directly predicting RGB pixel values in the masked region, the method first estimates the optical flow between adjacent frames.
Based on the estimated flow, pixel values from surrounding frames are then propagated to fill the masked regions.
By leveraging motion information, DFVI can accommodate complex object movements.  
However, it does not correct the propagation errors in pixel values in later processing modules.

To address this issue, Li \etal proposed E2FGVI~\cite{E2FGVI}, which is a video inpainting method inspired by optical flow based video super-resolution methods~\cite{BasicVSR,BasicVSRpp}.
Their method encodes input frames into feature maps instead of processing them in the image space, and uses downsampled optical flow to guide feature propagation between adjacent frames.
To enable accurate propagation of feature maps, E2FGVI employs DCN~\cite{DCNv1,DCNv2} for feature refinement.
Following BasicVSR++~\cite{BasicVSRpp}, this approach is motivated by the observation that offset diversity improves performance~\cite{understanding}.
However, even in regions where the optical flow estimation is accurate, the use of downsampled flow limits the achievable propagation accuracy.

To overcome this limitation, Zhou \etal proposed ProPainter~\cite{ProPainter}, which adaptively combines image space propagation and feature space propagation.
Their method first evaluates the validity of the optical flow using bidirectional consistency.
Only regions with high flow validity are processed in the image space, allowing precise pixel-level propagation.
This process partially fills the masked regions.
Subsequently, feature space propagation is applied to complete the remaining masked regions.
By integrating both propagation strategies, ProPainter effectively leverages accurate pixel information from regions with high flow validity via precise image propagation, while mitigating the adverse effects of flow errors through feature propagation.

As an alternative to optical flow-based approaches, Li \etal proposed a video inpainting method called DGDVI~\cite{DGDVI}, which utilizes depth maps as guidance.
Because optical flow may be unreliable in masked regions due to high temporal variability, they instead used depth maps as more reliable guidance, since depth maps tend to be more temporally stable.

As described above, various video inpainting methods have been proposed, but they are designed for conventional videos with a narrow FoV and do not address the distortion in omnidirectional videos.
Consequently, these methods cannot be considered optimal for omnidirectional video inpainting.

\noindent \textbf{Omnidirectional Inpainting.}
Several inpainting methods tailored for omnidirectional content have been proposed.
For example, Gkitas \etal~\cite{PanoDR} and Pintore \etal~\cite{Instant_Automatic} proposed omnidirectional image inpainting methods that incorporate padding strategies considering the horizontal cyclic continuity, as well as loss functions that take into account the distortion of omnidirectional images.
In addition, Gao \etal~\cite{Layout-Guided} proposed a method that preserves structural information of indoor scenes in omnidirectional images by leveraging a layout estimation model specifically designed for omnidirectional images.

Inpainting methods have been proposed not only for omnidirectional images but also for omnidirectional videos.
For example, Kawai \etal~\cite{2009kawai} proposed an omnidirectional video inpainting method to fill regions that the omnidirectional camera systems of that era could not capture.
Their method leverages structure-from-motion specifically adapted for omnidirectional videos to estimate 3D point clouds.
These point clouds are then used to identify regions in other frames that correspond to the missing region in the current frame, enabling the  inpainting process.
Kawai \etal~\cite{2014kawai} proposed a method that automatically removes dynamic objects from omnidirectional videos by leveraging structure-from-motion.
Xu \etal~\cite{2016ROBIO} observed that distortion in ERP increases toward the poles.
To address this, they rotate each frame so that the masked regions are positioned near the equator, where distortion is minimal, before performing optical flow based inpainting.
Choi \etal~\cite{OmniLocalRF} proposed a method based on Neural Radiance Fields (NeRF)~\cite{NeRF} for removing dynamic objects from omnidirectional videos.

However, all of these omnidirectional video inpainting methods assume that masked regions are visible in other frames.
In addition, the method by Kawai \etal~\cite{2009kawai} is limited to inpainting the ground region, while the method by Xu \etal~\cite{2016ROBIO} requires a static background.
These additional constraints limit their applicability.
Moreover, Kawai \etal~\cite{2014kawai} and Choi \etal~\cite{OmniLocalRF} focus on removing dynamic objects to recover a static scene and do not address the removal of static elements such as graffiti on walls.
In contrast, our method imposes no such constraints and enables inpainting of any user-specified region while preserving spatial and temporal consistency.

\section{Method}
\subsection{Overview}
\begin{figure}[tb]
  \centering
  \includegraphics[width=1.0\linewidth]{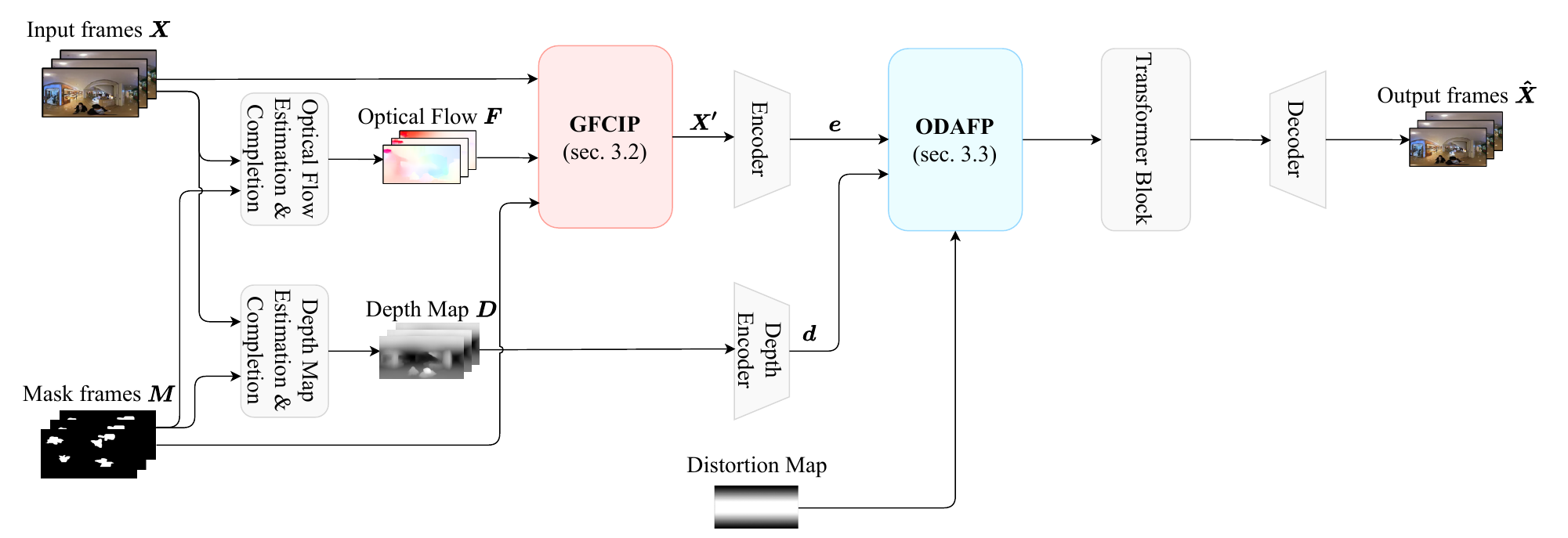}
  \caption{Overview of the proposed framework. It is divided into two main modules: the Geodesic Flow-Consistent Image Propagation (GFCIP) module and the Omnidirectional Depth-Assisted Feature Propagation (ODAFP) module.}
  \label{fig:my_method_all}
\end{figure}
This paper proposes Distortion-Aware Omnidirectional Video Inpainting (DAOVI), a deep learning model for video inpainting that accounts for distortion in omnidirectional videos.
The overall framework is shown in Fig.~\ref{fig:my_method_all}.
DAOVI takes as input the video frame sequence $\bm{X} = [X_1, X_2, \dots, X_T]$ and the corresponding mask sequence $\bm{M} = [M_1, M_2, \dots, M_T]$, where $T$ denotes the total number of frames.

To ensure spatial and temporal consistency, DAOVI uses optical flow, following high-performing deep learning-based video inpainting methods~\cite{DFVI, FGVC, E2FGVI, ProPainter}.
Optical flow $\bm{F} = \{F^f, F^b\}$ is estimated from the video frame sequence $\bm{X}$ and the mask sequence $\bm{M}$ using a pre-trained flow estimation model~\cite{RAFT} and a flow completion network~\cite{ProPainter}, where $F^f = \{F^f_t = F_{t \to t+1}\}^{T-1}_{t=1}$ and $F^b = \{F^b_t = F_{t+1 \to t}\}^{T-1}_{t=1}$ represent the forward and backward flows, respectively.

Next, the \emph{Geodesic Flow-Consistent Image Propagation (GFCIP)} module uses only reliable vectors in $\bm{F}$ to propagate pixel values from adjacent frames in the image domain. This module produces a partially inpainted output $\bm{X}'=[X'_1, X'_2, \dots, X'_T]$.
Note that the validity of the flow vectors is assessed by computing the geodesic distance to ensure suitability for omnidirectional videos.
Since regions with low flow validity remain unfilled, these areas are inpainted in the feature space by passing $\bm{X}'$ through a $9$‑layer CNN encoder to produce the feature maps $\bm{e}=[e_1, e_2, \dots, e_T]$.

Following GFCIP, the Omnidirectional Depth-Assisted Feature Propagation (ODAFP) module propagates information from adjacent frames in the feature space.
Specifically, ODAFP employs DCN~\cite{DCNv1, DCNv2} to propagate features from adjacent feature maps, guided by optical flow.
To avoid relying exclusively on optical flow, ODAFP also incorporates depth information.
The depth map $\bm{D} = [D_1, D_2, \dots, D_T]$ is estimated from the video frame sequence $\bm{X}$ and the mask sequence $\bm{M}$ via a pre-trained omnidirectional depth estimation model~\cite{depthanywhere} and a depth completion network~\cite{DGDVI}.
The parameters of both networks remain fixed during DAOVI training.
To integrate depth information into feature space, a 2-layer CNN encoder generates depth feature maps $\bm{d} = [d_1, d_2, \dots, d_T]$.
Furthermore, to account for the geometric distortion inherent in omnidirectional videos during propagation, DCN offsets and modulation masks are weighted by a distortion map~\cite{WS-PSNR}, which represents per-pixel ERP distortion.

We incorporate a transformer-based module to further enhance the model’s expressive capacity.  
Specifically, we adopt the Mask-Guided Sparse Video Transformer~\cite{ProPainter}, a transformer-based module that is computationally efficient.
Finally, a $4$‑layer CNN decoder reconstructs the fully inpainted video frames $\hat{\bm{X}}=[\hat X_1,\hat X_2,\dots,\hat X_T]$.
The following sections describe the two core modules (GFCIP and ODAFP) in detail.

\subsection{Geodesic Flow-Consistent Image Propagation (GFCIP)}
\label{sec:img_prop}
\begin{figure}
\centering
\setlength{\tabcolsep}{8pt}
\begin{tabular}{cc}
\bmvaHangBox{\includegraphics[height=2.5cm]{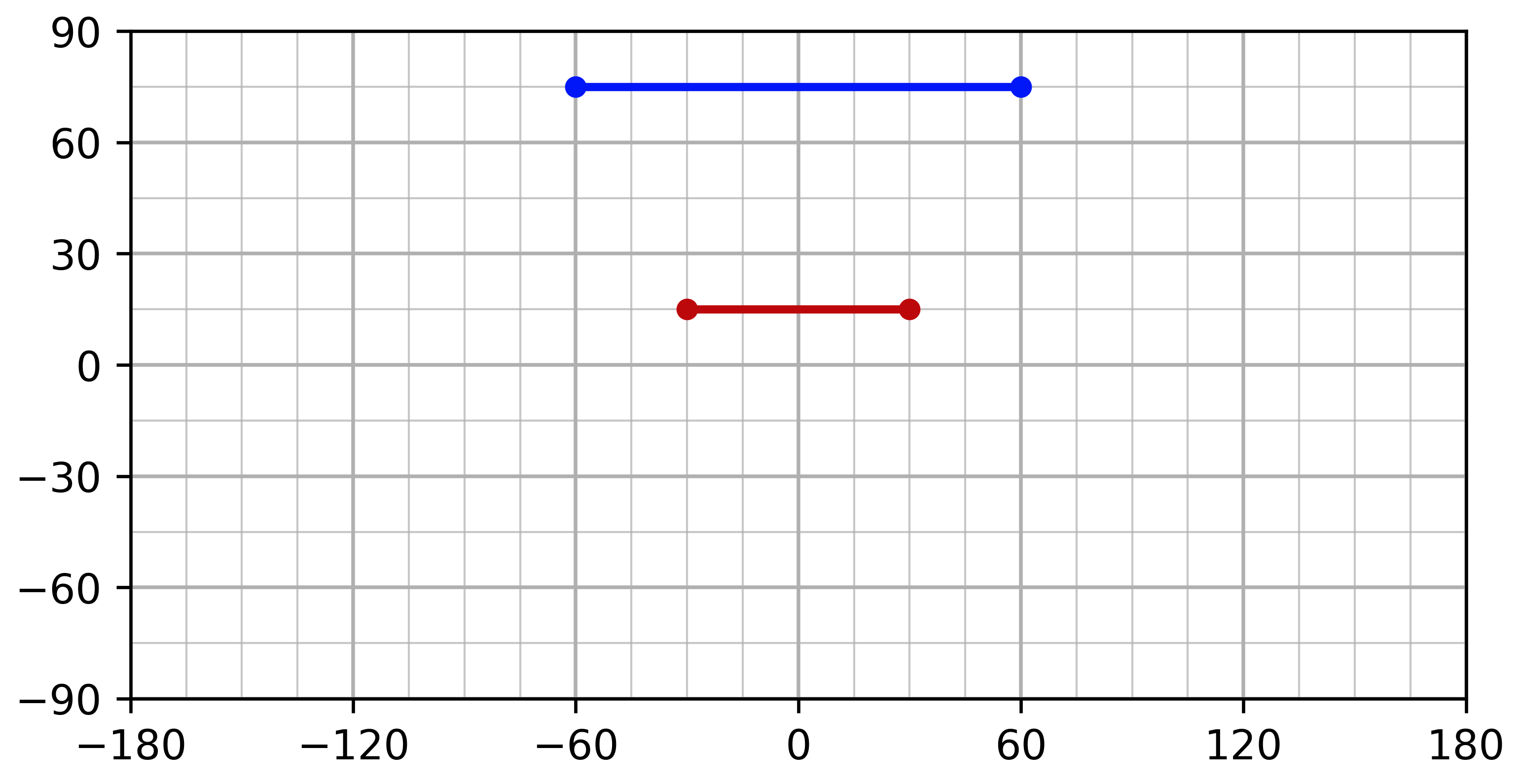}} &
\bmvaHangBox{\includegraphics[height=2.5cm]{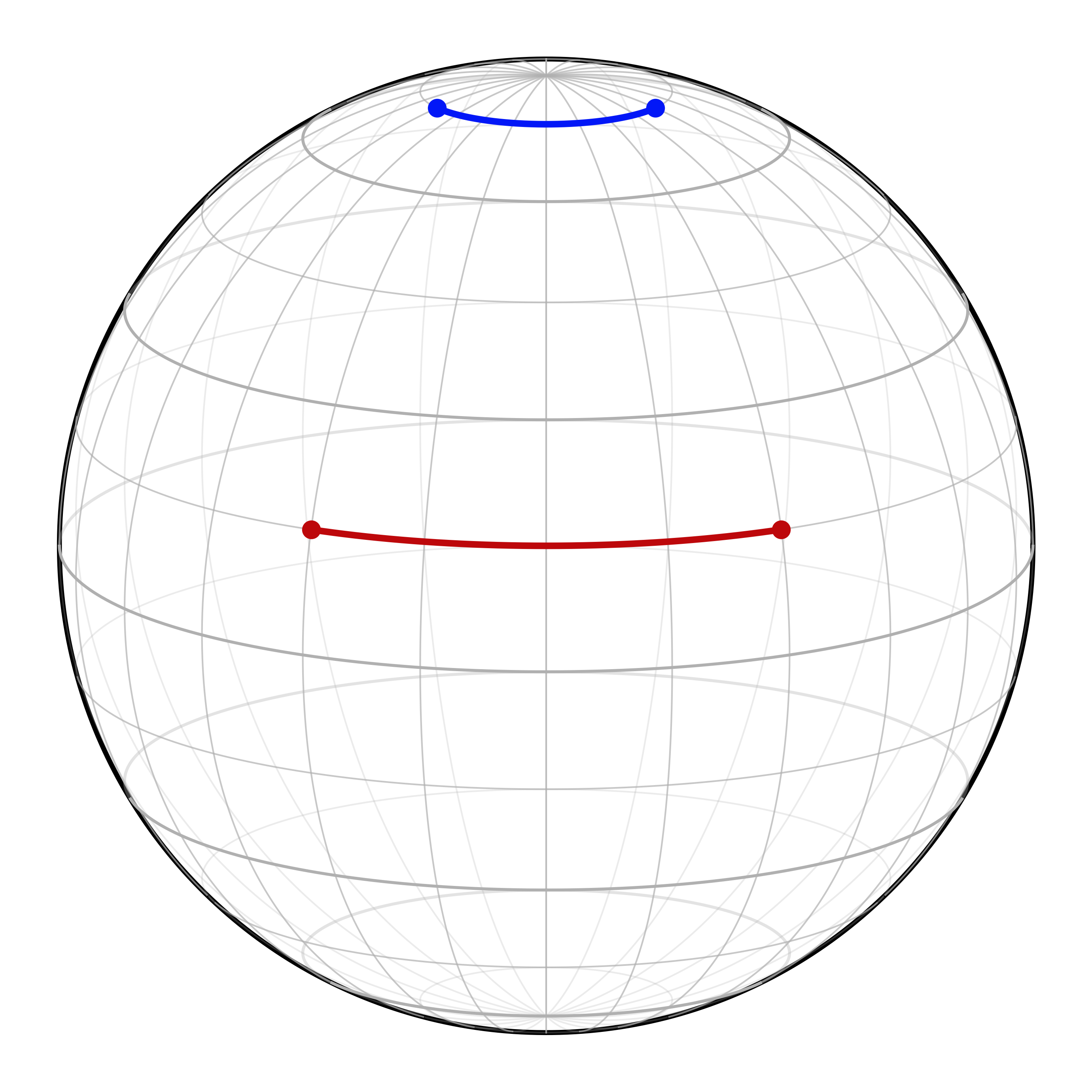}} \\
\end{tabular}
\caption{Left: two line segments shown in ERP pixel coordinates. The blue segment appears longer than the red segment. Right: the actual lengths of the two line segments. The red segment is actually longer than the blue segment.}
\label{fig:ERP_sphere}
\end{figure}

The output $X'_t$ of the GFCIP module is computed from the input frame $X_t$, its next frame $X_{t+1}$, and the optical flow $F_{t\to t+1}$ representing the motion between frame $X_t$ and $X_{t+1}$.
This output can be formulated as
\begin{equation}
X'_t = \mathcal{W}(X_{t+1}, F_{t\to t+1}) * M_r + X_t * \bigl(1 - M_r\bigr)
\end{equation}
where $\mathcal{W}(\cdot)$ is the warping operation.
The binary mask $M_r$ identifies those pixels within the masked region that satisfy the required flow validity for reliable propagation.
For a pixel position $p$, the binary mask $M_r(p)$ is defined as
\begin{equation}
\label{eq:valid_area}
M_r(p) = 
\begin{cases}
1 & \text{if } p \in C_1 \cap C_2 \cap C_3 \\
0 & \text{otherwise.}
\end{cases}
\end{equation}
The conditions $C_1$, $C_2$, and $C_3$ are defined as follows.
Condition $C_1$ tests whether the optical flow at a given pixel is sufficiently reliable.
Since the flow is estimated by a pre-trained model and may contain errors, this check is required.
The position $p'$ is obtained by first mapping a pixel at $p$ from frame $X_t$ to $X_{t+1}$ via the optical flow $F_{t\to t+1}$, and then mapping it back from frame $X_{t+1}$ to $X_t$ via $F_{t+1\to t}$.
It can be expressed as
\begin{equation}
\label{eq:new_position}
p' = p + F_{t\to t+1}(p) + F_{t+1\to t}\bigl(p + F_{t\to t+1}(p)\bigr)
\end{equation}
If the optical flow were perfectly accurate, those two mappings would cancel out, and $p'$ would coincide with $p$.
Accordingly, the distance between $p$ and $p'$ quantifies the magnitude of the flow consistency error.
As shown in Fig.~\ref{fig:ERP_sphere}, Euclidean distance in ERP pixel coordinates does not coincide with the actual distance.
Therefore, this distance is evaluated using geodesic distance.
We denote this geodesic distance by $E(\cdot)$ and compute it as
\begin{equation}
E(a, b) = \arccos\bigl(\cos\theta_a\,\cos\theta_b\,\cos(\phi_a-\phi_b) + \sin\theta_a\,\sin\theta_b\bigr)
\end{equation}
\begin{align}
\phi_p &= \frac{2\pi\,(x_p + 0.5)}{W} \;-\;\pi
\quad,\quad p \in \{a, b\} \\
\theta_p &= \frac{\pi\,(y_p + 0.5)}{H}
\quad,\quad p \in \{a, b\}
\end{align}
where $\phi$ and $\theta$ are the spherical coordinates and $W$ and $H$ denote the width and height of the input video frames, respectively.
Using this notation, $C_1$ is formalized as
\begin{equation}
C_1: \quad E\bigl(p,\,p'\bigr) < \epsilon
\end{equation}
where $\epsilon$ is the threshold, which we set to $0.4^\circ$.

Condition $C_2$ signifies that the pixel position $p$ in frame $X_t$ is located in the masked region, which can be expressed as  
\begin{equation}
C_2:M_t(p) = 1    
\end{equation}
This condition ensures that the inpainting target region $M_r$ is a subset of the mask.

Condition $C_3$ requires that the position reached by moving $p$ according to the optical flow $F_{t \to t+1}$ lies outside the mask in frame $t+1$, which can be expressed as  
\begin{equation}
C_3: M_{t+1}\bigl(p + F_{t\to t+1}(p)\bigr) = 0
\end{equation}
This condition ensures that the source location for propagation is unmasked.

\begin{figure}[htb]
\begin{center}
\includegraphics[width=1.0\linewidth]{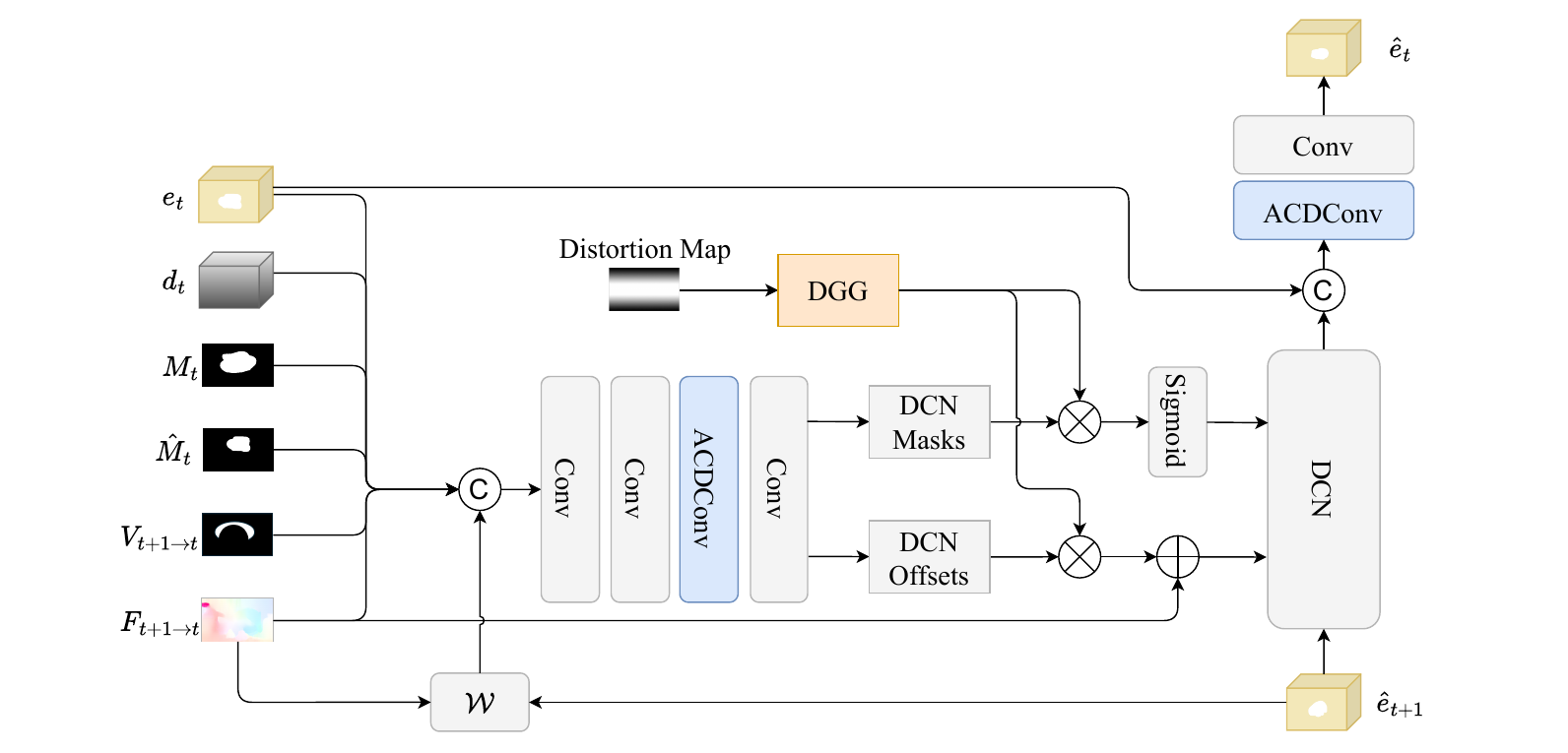}
\caption{Omnidirectional Depth-Assisted Feature Propagation (ODAFP). 
It leverages optical flow and depth feature maps for propagation in feature space, and uses a distortion map to weight DCN offsets and modulation masks.}
\label{fig:ODAFP}
\end{center}
\end{figure}

\subsection{Omnidirectional Depth-Assisted Feature Propagation (ODAFP)}
\label{sec:feat_prop}
Fig.~\ref{fig:ODAFP} shows the ODAFP module.
ODAFP propagates features from adjacent frames in the feature space using DCN~\cite{DCNv1, DCNv2} and optical flow, following high-performing existing video inpainting approaches~\cite{E2FGVI, ProPainter}.
Specifically, ODAFP first aligns the adjacent frame output $\hat{e}_{t+1}$ to the current frame $t$ using a DCN guided by the optical flow $F_{t+1 \to t}$.
The aligned feature map and current frame feature map $e_t$ are then concatenated and passed through convolutional layers to produce the output of this module $\hat{e}_t$.
Similarly, feature propagation using the forward optical flow $F_{t\to t+1}$ is performed.
For simplicity, the following description focuses on propagation with the backward optical flow $F_{t+1\to t}$.

DCN requires offsets and modulation masks.
To generate these, the following components are concatenated and fed through convolutional layers: the current feature map $e_t$, the mask $M_t$, the unfilled region mask $\hat{M}_t$ from GFCIP, backward optical flow $F_{t+1 \to t}$, its validity map $V_{t+1 \to t}$, the warped adjacent frame output $\mathcal{W}\bigl(\hat{e}_{t+1}\bigr)$, and the depth feature map $d_t$.
Including $d_t$ avoids relying solely on optical flow and strengthens robustness.

The concatenated features are then processed by a sequence of 4 convolutional layers.
In one of these layers, ODAFP uses the Adaptively Combined Dilated Convolution (ACDConv)~\cite{ACDNet}, which is specifically designed for omnidirectional images.
ACDConv dynamically weights and combines the outputs of multiple dilated convolution layers~\cite{dilated_conv}, which enables the operation to adapt to the spatially varying distortions in omnidirectional videos.

Additionally, instead of standard zero padding, circular padding~\cite{CircPad} is applied in the convolutional layers.
This padding method preserves the cyclic continuity between the leftmost and rightmost columns, and also preserves continuity at the poles in ERP images.

\begin{figure}[tb]
\begin{center}
\includegraphics[width=0.3\linewidth]{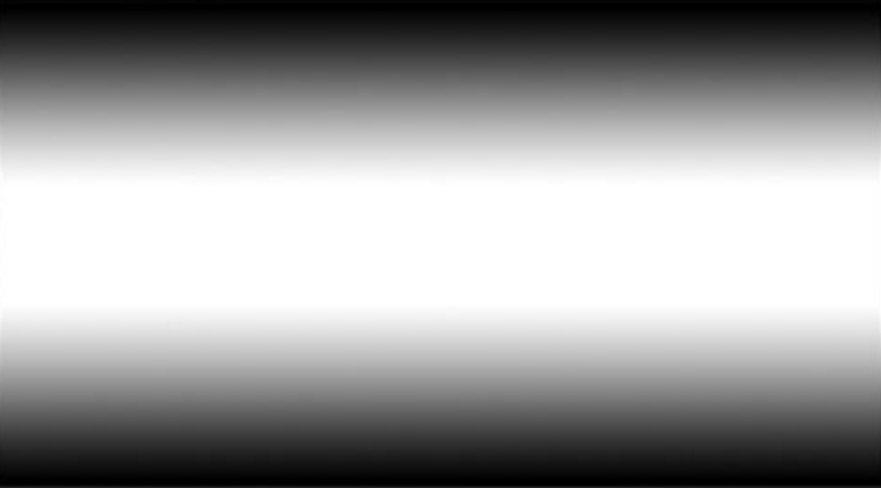}
\caption{Distortion Map. Distortion due to the equirectangular projection is lower in brighter regions and higher in darker regions. The distortion magnitude depends solely on the y-coordinate.}
\label{fig:distortion_map}
\end{center}
\end{figure}

Moreover, to better adapt the offsets and modulation masks to omnidirectional videos, we utilize the ERP distortion weighting formula~\cite{WS-PSNR}.
The per-pixel ERP distortion weight $w_{\mathrm{erp}}(i,j)$ at coordinate $(i,j)$ is defined as
\begin{equation}
\label{eq:distortion}
w_{\mathrm{erp}}(i,j) = \cos\!\biggl(\frac{(j + 0.5 - N/2)\,\pi}{N}\biggr)
\end{equation}
where $N$ denotes the height of the ERP image.
As shown in Equation~\ref{eq:distortion}, the ERP distortion magnitude $w_{\mathrm{erp}}(i,j)$ is determined solely by the pixel's y-coordinate $j$.
A per-pixel distortion map is constructed by applying $w_{\mathrm{erp}}(i,j)$ at each coordinate.
As shown in Fig.~\ref{fig:distortion_map}, the per-pixel distortion weight $w_{\mathrm{erp}}$ is larger near the equator and smaller toward the poles, taking values close to $1$ around the equator and tapering toward $0$ near the poles.
This indicates that distortion due to the equirectangular projection is minimal near the equator and increases toward the poles.

To effectively utilize the distortion map, we employ the Distortion Guidance Generator (DGG) module~\cite{GDGT-OSR}.
DGG encodes the per-pixel distortion map into the latent space and produces distortion guidance $G$.  
This distortion guidance $G$ is used to weight the DCN offsets and modulation masks.
The weighted offsets are then added to the optical flow $F_{t+1 \to t}$, and the weighted modulation masks are passed through a sigmoid function before being applied in the DCN.

\section{Experiments}
In this section, we describe the experiments conducted to validate the effectiveness of the proposed method.
Implementation details, additional ablation study and additional qualitative results are provided in the supplementary material.

\subsection{Experimental Settings}
\noindent \textbf{Dataset.} 
In this study, we used the ODV360 omnidirectional video dataset~\cite{ODV360}.
ODV360 contains 210 videos for training, 20 for validation, and 20 for testing, with each video consisting of 100 frames.
We adhered to this split for model training and evaluation.
Although the original videos in ODV360 have a resolution of $540\times270$ pixels, we downsampled them to $304\times152$ pixels due to computational resource constraints.

Since ODV360 does not provide mask frames for the video inpainting task, we generated random mask sequences for evaluation, as used for quantitative evaluation in existing video inpainting methods~\cite{E2FGVI, ProPainter}.
Unlike the random masks used in prior video inpainting methods, which are typically single and static, our evaluation uses sequences composed of multiple randomly shaped regions that move independently across frames.
This evaluation setting spatially disperses mask positions because distortion varies with coordinates in omnidirectional videos, as noted in Section~\ref{sec:feat_prop}, in order to ensure a fair comparison.
The same random masks are used for qualitative evaluation.

\noindent \textbf{Network and Training.} 
We train our model using the Adam~\cite{Adam} optimizer with a batch size of 8, setting the learning rate to $1.5\times10^{-4}$ and running $80$k iterations.
Our implementation is based on the PyTorch framework, and training is performed on a single NVIDIA RTX 6000 Ada GPU.

\noindent \textbf{Evaluation Metrics.} 
For the quantitative evaluation of inpainting results, we employed five metrics: PSNR, SSIM~\cite{SSIM}, WS-PSNR~\cite{WS-PSNR}, WS-SSIM~\cite{WS-SSIM}, and VFID~\cite{VFID}.
WS-PSNR and WS-SSIM are weighted versions of PSNR and SSIM~\cite{SSIM}, respectively, which account for the positional distortion inherent in omnidirectional videos.
VFID is an extension of the image quality metric FID~\cite{FID} to videos, and it evaluates perceptual video quality similarity using a pre‑trained CNN model~\cite{I3D,ResNeXt}.

\begin{table}[tb]
\caption{Quantitative comparison on ODV360~\cite{ODV360} dataset. The best result of each metric is marked in \textbf{bold} font.}
\label{tab:quantitative}
\centering
\begin{tabular}{c|ccccc}
\hline
    Method & PSNR[dB]($\uparrow$) & SSIM($\uparrow$) & WS-PSNR[dB]($\uparrow$) & WS-SSIM($\uparrow$) & VFID($\downarrow$)\\ \hline
    FuseFormer & 29.18 & 0.9727 & 28.87 & 0.9392 & 0.320\\
    STTN & 32.21 & 0.9815 & 31.09 & 0.9520 & 0.238\\
    ProPainter & 32.81 & 0.9877 & 32.86 & 0.9648 & 0.149\\
    Ours & \textbf{33.37} & \textbf{0.9888} & \textbf{33.37} & \textbf{0.9661} & \textbf{0.138}\\ \hline
\end{tabular}
\end{table}

\subsection{Comparison with Baseline Methods}
\noindent \textbf{Quantitative Results.}
We report quantitative results on the ODV360 dataset~\cite{ODV360}.
In this experiment, we compared our method with several state-of-the-art (SOTA) video inpainting methods, including FuseFormer~\cite{FuseFormer}, STTN~\cite{STTN}, and ProPainter~\cite{ProPainter}, to evaluate the performance of our method.
As shown in Tab.~\ref{tab:quantitative}, our DAOVI surpasses existing video inpainting methods on all evaluation metrics.
The high PSNR and SSIM scores indicate high reconstruction fidelity, while the low VFID score demonstrates perceptually plausible inpainted frames.
Moreover, the high WS-PSNR and WS-SSIM confirm that DAOVI exhibits excellent performance for omnidirectional videos.

\begin{figure}[htbp]
  \centering
  \setlength{\tabcolsep}{1pt}
  \begin{tabular}{@{}ccccc@{}}
    \bmvaHangBox{\includegraphics[width=0.19\linewidth]{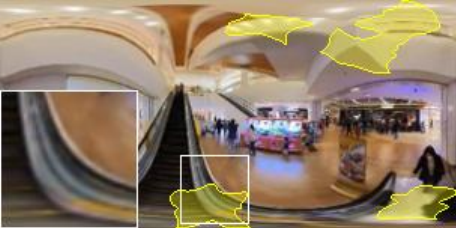}} &
    \bmvaHangBox{\includegraphics[width=0.19\linewidth]{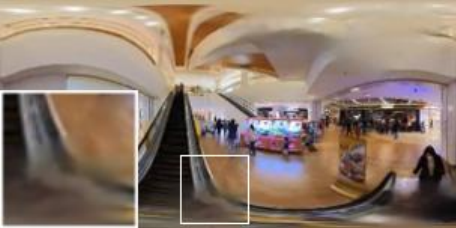}} &
    \bmvaHangBox{\includegraphics[width=0.19\linewidth]{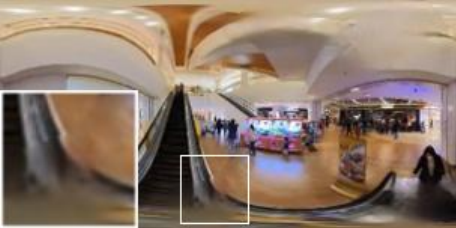}} &
    \bmvaHangBox{\includegraphics[width=0.19\linewidth]{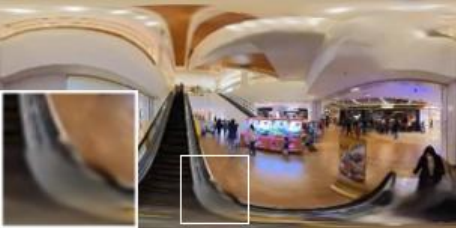}} &
    \bmvaHangBox{\includegraphics[width=0.19\linewidth]{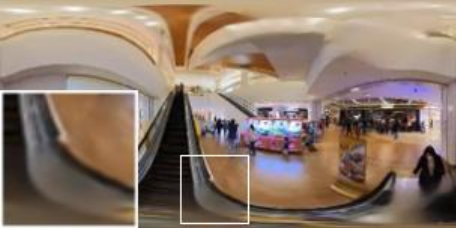}} \\
    
    \bmvaHangBox{\includegraphics[width=0.19\linewidth]{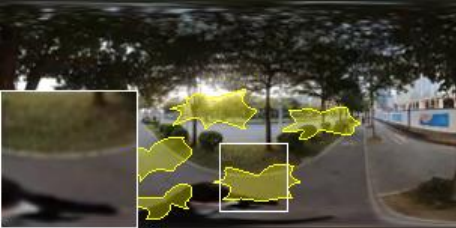}} &
    \bmvaHangBox{\includegraphics[width=0.19\linewidth]{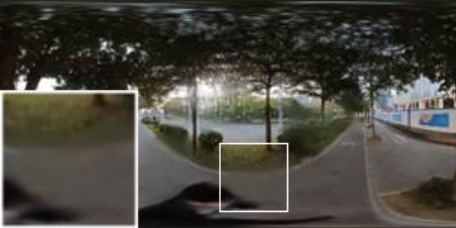}} &
    \bmvaHangBox{\includegraphics[width=0.19\linewidth]{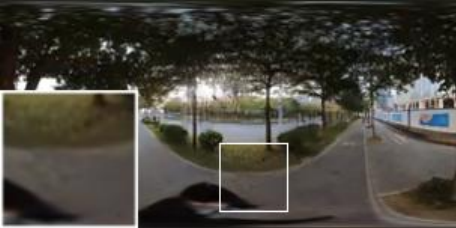}} &
    \bmvaHangBox{\includegraphics[width=0.19\linewidth]{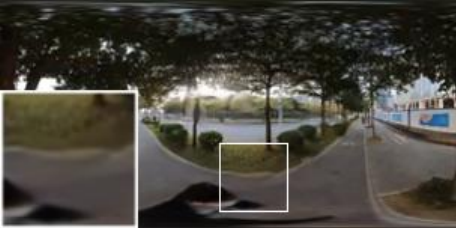}} &
    \bmvaHangBox{\includegraphics[width=0.19\linewidth]{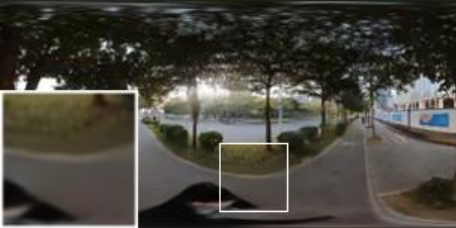}} \\

    \bmvaHangBox{\includegraphics[width=0.19\linewidth]{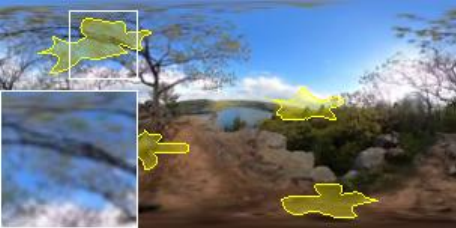}} &
    \bmvaHangBox{\includegraphics[width=0.19\linewidth]{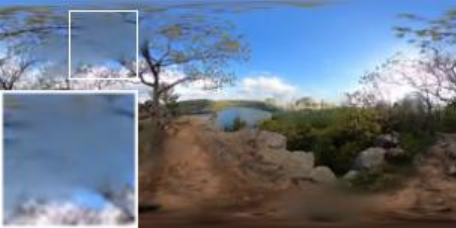}} &
    \bmvaHangBox{\includegraphics[width=0.19\linewidth]{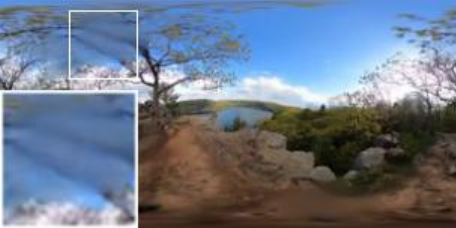}} &
    \bmvaHangBox{\includegraphics[width=0.19\linewidth]{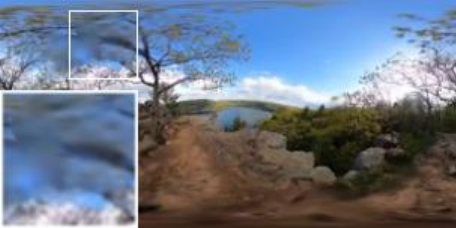}} &
    \bmvaHangBox{\includegraphics[width=0.19\linewidth]{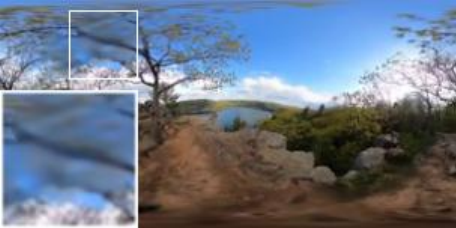}} \\

    \bmvaHangBox{\includegraphics[width=0.19\linewidth]{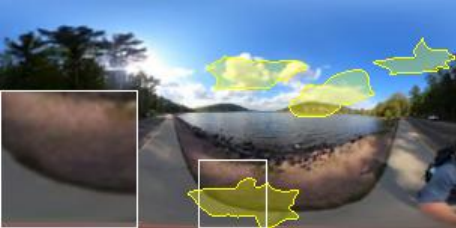}} &
    \bmvaHangBox{\includegraphics[width=0.19\linewidth]{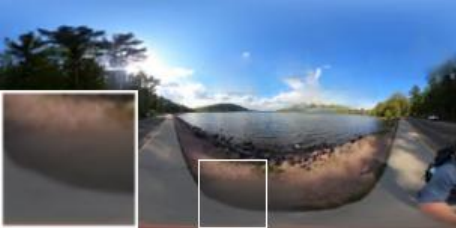}} &
    \bmvaHangBox{\includegraphics[width=0.19\linewidth]{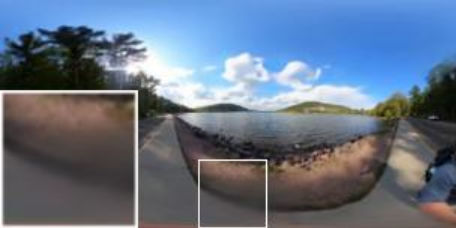}} &
    \bmvaHangBox{\includegraphics[width=0.19\linewidth]{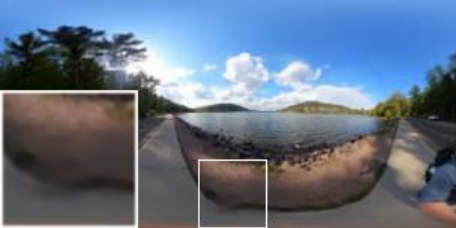}} &
    \bmvaHangBox{\includegraphics[width=0.19\linewidth]{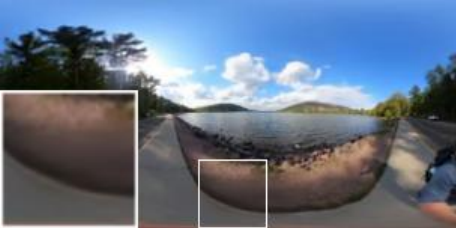}} \\
    Masked Frames & FuseFormer & STTN & ProPainter & DAOVI (Ours)
  \end{tabular}
  \caption{Inpainting results of the proposed DAOVI and several SOTA video inpainting methods~\cite{FuseFormer, STTN, ProPainter} originally designed for ordinary videos with a narrow FoV. The yellow regions indicate the masked areas, and the outputs of each model within these regions are enlarged and shown at the bottom left. By explicitly handling geometric distortion, our method produces qualitatively better inpainting with improved structural consistency and fewer artifacts.}
  \label{fig:qualitative}
\end{figure}

\noindent \textbf{Qualitative Results.}
For the visual comparison, we compare our method with other video inpainting approaches, including FuseFormer~\cite{FuseFormer}, STTN~\cite{STTN}, and ProPainter~\cite{ProPainter}.
As shown in Fig.~\ref{fig:qualitative}, applying methods designed for narrow-FoV videos to omnidirectional inputs produces noticeable artifacts and unsatisfactory reconstructions.
In contrast, our DAOVI produces visually plausible results.
This demonstrates the effectiveness of the proposed method for omnidirectional video inpainting.

\subsection{Ablation Study}
\begin{table}[tb]
\centering
\caption{Ablation Study. The best result of each metric is marked in \textbf{bold} font.}
\label{tab:ablation}
\begin{tabular}{cccccc}
\hline
    Method & PSNR[dB]($\uparrow$) & SSIM($\uparrow$) & WS-PSNR[dB]($\uparrow$) & WS-SSIM($\uparrow$) & VFID($\downarrow$)\\ \hline
    w/o GFCIP & 33.25 & 0.9884 & 33.25 & 0.9656 & \textbf{0.136} \\
    w/o ODAFP & 32.84 & 0.9878 & 32.89 & 0.9650 & 0.148 \\
    Ours & \textbf{33.37} & \textbf{0.9888} & \textbf{33.37} & \textbf{0.9661} & 0.138\\
    \hline
\end{tabular}
\end{table}
An ablation study is conducted to verify the effectiveness of the proposed modules.
As shown in Tab.~\ref{tab:ablation}, removing either GFCIP or ODAFP leads to degraded scores relative to the full model.
These results indicate that both modules enhance omnidirectional video inpainting performance.
The larger score drop when ODAFP is removed, which highlights its stronger contribution.
By contrast, GFCIP contributes less to performance improvement than ODAFP, but can be incorporated without introducing any additional learnable parameters.

\section{Conclusion}
In this work, we propose a deep learning–based omnidirectional video inpainting framework called DAOVI.
The proposed framework includes modules that perform distortion-aware propagation in both the image space and the feature space to address the unique geometry of omnidirectional videos.
In the image space, pixel values are propagated only along reliable optical flow.
Reliability of optical flow is assessed by the consistency error between forward and backward flows, measured in geodesic distance rather than Euclidean distance in ERP pixel coordinates.
In the feature space, to adapt to ERP specific distortion, distortion aware convolution is employed.
In addition, intermediate features of the model are weighted with a distortion map that encodes the per pixel magnitude of ERP distortion.
Furthermore, to avoid excessive reliance on the accuracy of the estimated optical flow, the estimated depth map is also used as an input to the model.
Experimental results demonstrate that DAOVI outperforms SOTA video inpainting methods designed for ordinary videos with a narrow field of view in both quantitative and qualitative evaluations.

\medskip
\noindent \textbf{Acknowledgment.}
This work was partially supported by JSPS KAKENHI(A) 25H01159.

\newpage
\bibliography{arXiv_egbib}
\end{document}